%% file: main.tex
\documentclass[10pt]{article} %
\usepackage[table,dvipsnames]{xcolor}
\usepackage[preprint]{tmlr}

\usepackage{hyperref}
\usepackage{url}

\definecolor{LightAquaGreen}{RGB}{212, 252, 223}
\usepackage{graphicx} %
\usepackage{caption}
\usepackage{subcaption}
\usepackage{wrapfig}

\usepackage{booktabs}
\usepackage{multirow}
\usepackage{multicol}
\usepackage{microtype}
\usepackage{makecell}

\usepackage{siunitx}
\usepackage{enumitem}
\usepackage{algorithm}
\usepackage{algpseudocode}
\usepackage{bbm}
\usepackage{float}

\usepackage{amsmath}
\usepackage{amsfonts}
\usepackage{amssymb}
\usepackage{amsthm}

\input{aliases}

\title{Inverting Gradient Attacks Makes Powerful Data Poisoning} %

\author{\name Wassim (Wes) Bouaziz \email wesbz@meta.com \\
      \addr Meta, FAIR \& CMAP, École polytechnique
      \AND
      \name El-Mahdi El-Mhamdi \\
      \addr CMAP, École polytechnique
      \AND
      \name Nicolas Usunier \\
      \addr Work done at Meta, FAIR}

\def\month{MM}  %
\def\year{YYYY} %
\def\openreview{\url{https://openreview.net/forum?id=XXXX}} %

\begin{document}

\maketitle

\input{sections/00-abstract}
\input{sections/01-introduction}
\input{sections/02-background}
\input{sections/03-framework}
\input{sections/04-method}
\input{sections/05-experiments}
\input{sections/06-discussion}
\input{sections/07-conclusion}

\bibliography{main}
\bibliographystyle{tmlr}

\newpage
\appendix
\input{sections/08-appendix}

\end{document}

%% file: aliases.tex
 \newcommand{\nico}[1]{{\color{blue} N:}}

%% file: sections/00-abstract.tex
\begin{abstract}
Gradient attacks and data poisoning tamper with the training of machine learning algorithms to maliciously alter them and have been proven to be equivalent in convex settings.
The extent of harm these attacks can produce in non-convex settings is still to be determined.
Gradient attacks can affect far less systems than data poisoning but have been argued to be more harmful since they can be arbitrary, whereas data poisoning reduces the attacker's power to only being able to inject data points to training sets, via e.g. legitimate participation in a collaborative dataset.
This raises the question of whether the harm made by gradient attacks can be matched by data poisoning in non-convex settings.
In this work, we provide a positive answer in a worst-case scenario and show how data poisoning can mimic a gradient attack to perform an availability attack on (non-convex) neural networks.
Through gradient inversion, commonly used to reconstruct data points from actual gradients, we show how reconstructing data points out of malicious gradients can be sufficient to perform a range of attacks.
This allows us to show, for the first time, an availability attack on neural networks through data poisoning, that degrades the model's performances to random-level through a minority (as low as 1\%) of poisoned points.
\end{abstract}

%% file: sections/01-introduction.tex
\section{Introduction}

Security in Machine Learning has come to consider various attackers with a wide range of capabilities.
In training time attacks, an attacker may only need to participate in the training procedure and send strategically chosen, yet legitimate-looking participation.
We call \emph{gradient attacks} the cases where the attacker can send gradients, and \emph{data poisoning} when they send data points.
The severity of attacks range from integrity attacks, where the model's reliability or consistency can be altered at different levels of severity, to the most severe, availability attacks, where the model is plainly unusable, e.g. due to performances being too low to allow for normal operation.

So far, unless the objective function is convex \cite{biggio2013poisoning}, availability attacks have only been possible through gradient attacks \cite{blanchard2017byzantinetolerant, mhamdi2018hidden, baruch2019little}, leaving an open question of whether gradient attacks are fundamentally more powerful than data poisoning attacks outside the convex setting.
In this work, we answer this question by providing empirical evidence that poisoning attacks can lead to availability attacks on (non-convex) neural networks, under a threat model that allows for comparing both gradient and data poisoning attacks.

\begin{figure}[ht]
    \centering
    \includegraphics[width=0.8\linewidth]{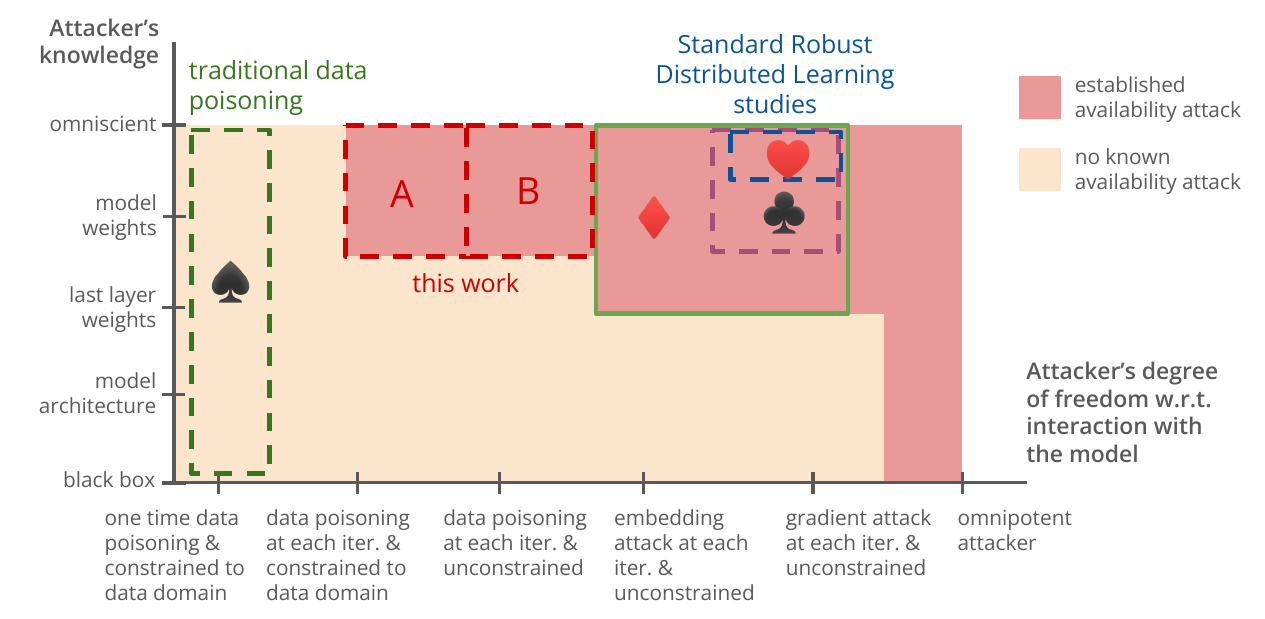}
    \caption{Territory of known availability attacks (in red) within a domain of constraints. The closer to the origin, the more constrained is the setting for the attacker and the harder it is to realize an availability attack.
    $\spadesuit$: \cite{geiping2021witches, zhao_clpa_2022,ning2021invisible, huang2020metapoison}, 
    $\heartsuit$: \cite{blanchard2017byzantinetolerant,baruch2019little}, 
    $\clubsuit$: \cite{mhamdi2018hidden},
    $\diamondsuit$ so far only in convex settings : \cite{farhadkhani2022equivalence},
    A: Result A in subsection \ref{subsec:results},
    B: Result B in subsection \ref{subsec:results}.
    }
    \label{fig:this-work}
\end{figure}

\begin{figure}[ht]
    \centering
    \includegraphics[width=0.4\linewidth]{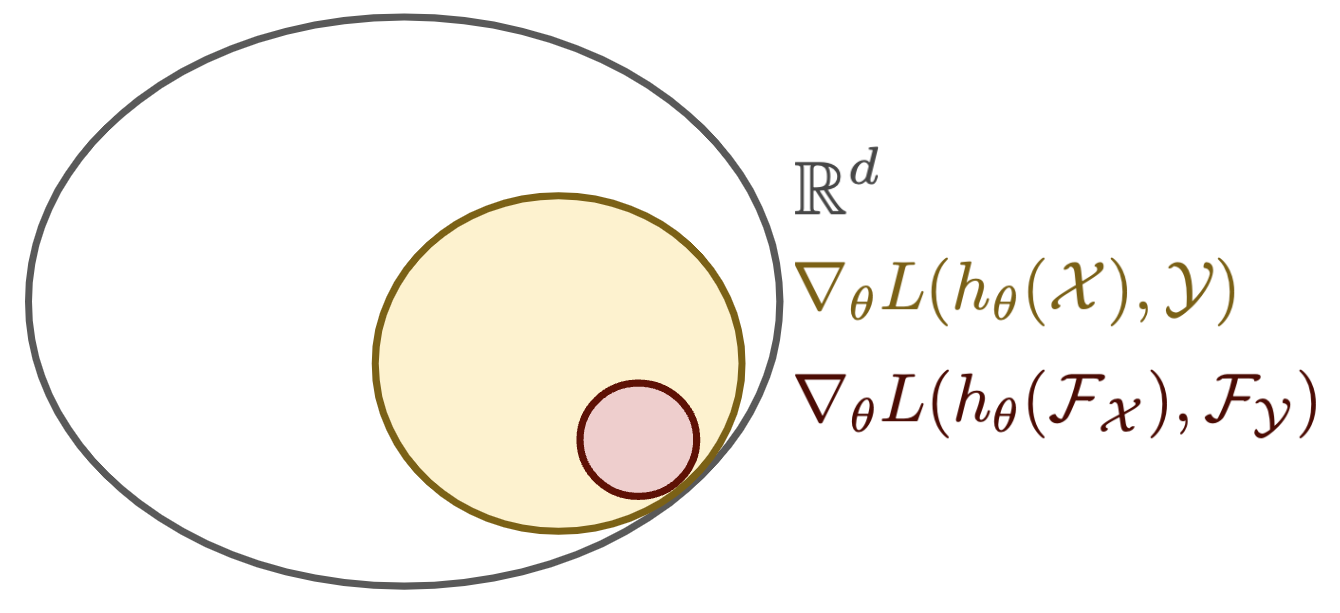}
    \caption{Images of the gradient operator on different sets. $\mathbb{R}^d$ is where an attacker can craft unrestricted gradient attacks. $\nabla_{\theta} L(h_{\theta}(\mathcal{X}), \mathcal{Y})$ is the set of possible gradients given an unrestricted data poisoning (Result B in subsection \ref{subsec:results}), and $\nabla_{\theta} L(h_{\theta}(\mathcal{F}_{\mathcal{X}}), \mathcal{F}_{\mathcal{Y}})$ is the set of possible gradients when data poisoning is restricted to a feasible set $\mathcal{F}_{\mathcal{X}} \times \mathcal{F}_{\mathcal{Y}} \subseteq \mathcal{X} \times \mathcal{Y}$ (Result A in subsection \ref{subsec:results}).}
    \label{fig:image-grad}
\end{figure}

An ``apple-to-apple'' comparison of data poisoning attacks and gradient attacks is not straightforward because the literature associates them with different threat models offering different levels of knowledge and interventions.
For instance, gradient attacks can be crafted at each iteration, while data poisoning attacks are usually crafted once and for all.
In order to remove this confounding factor, we consider a threat model in which \textbf{both attacks} can be executed by allowing an attacker to recalculate its attack at each iteration, similarly to Algorithm~1 in \citet{steinhardt2017certified}, may it be gradients or data poisoning (Figure \ref{fig:this-work}).
Once this confounding factor removed, the fundamental difference between gradient and data poisoning attacks comes from the limited expressivity of data poisoning compared to the arbitrary gradient attacks.

Our approach leverages gradient inversion methods, which were previously used in privacy attacks in distributed learning to reconstruct training data points from actual gradients they induced.
Contrary to privacy attacks, malicious gradients might not be achievable by the gradient operator if participants are expected to send legitimate-looking inputs (Figure \ref{fig:image-grad}).
We reconstruct data points whose induced gradient can replicate as much as possible a malicious gradient, and show that they can constitute a sufficient data poisoning to achieve an availability attack.

In our experiments, we exhibit a successful availability attack on recent neural network architectures, trained on an image classification task with different optimization algorithms, even when protected by a state-of-the-art defense mechanism against gradient attacks.
We show that: (1) the additional constraints under which data poisoning attacks operate, compared to gradient attacks, make them overall less effective than plain gradient attacks, and (2) the severity of data poisoning attacks covers the same range as gradient attacks, including availability attacks, even on non-convex neural networks.

Our contributions are the following:
\begin{itemize}
    \item We leverage gradient inversion mechanism from privacy attacks to reconstruct data poisoning from existing gradient attacks;
    \item We experiment this attack on neural networks under various training settings;
    \item We exhibit the first data poisoning availability attack on neural networks.
\end{itemize}

%% file: sections/02-background.tex
\section{Background}
\paragraph{Gradient attacks} have long been studied in the standard Robust Distributed Learning literature, %
when an attacker can send \textbf{arbitrary}\footnote{Often called Byzantine attacks, in reference to the Byzantine faults model in distributed computing~\cite{lamportByzGen}} gradients.
Lemma 1 in \cite{blanchard2017byzantinetolerant} shows that with stochastic gradient descent ($\textsc{SGD}$), availability attacks are possible with only a single malicious worker.
The attacker's freedom allows for a variety of attacks by exploiting the geometry of honest gradients and the inevitable weaknesses\cite{mhamdi2018hidden, baruch2019little, xie2019fall, Shejwalkar2021ManipulatingTB} of robust gradient aggregators when the model is high-dimensional or when the data is heterogeneous.
The attacker's goal is to bring the average gradient to another direction, making a step toward their objective.
There is no naive way to determine if a gradient is legitimate or not.
In such settings, defending is often considered as a task of robust mean estimation.
Robust aggregation of gradients is hence a possible defense mechanisms one can deploy \cite{blanchard2017byzantinetolerant, mhamdi2018hidden, yin2021byzantinerobust}.
Still, with better defense mechanisms came better attacks \cite{baruch2019little}.
Even stronger, Theorem~2 in \cite{elmhamdi2023impossible} shows the impossibility of robust mean estimation below a certain threshold that grows with data heterogeneity and model size.
Our work relies on these gradient attacks and shows how they can be transferred to data poisoning as successful availability attacks, even in non-convex settings.

\paragraph{Data poisoning} is the manipulation of training data of ML algorithms with the goal of influencing the algorithms' behavior. Several approaches exist to generate poisons: label flipping \cite{Shejwalkar2021BackTT}, generative methods \cite{munozgonzalez2019poisoning, zhao_clpa_2022}, and gradient-based approaches \cite{munozgonzalez2017poisoning, shafahi2018poison, geiping2021witches}.
The last allows to finely control the resulting gradient on the poisonous points instead of relying on another proxy.
Although \cite{Shejwalkar2021BackTT, Shejwalkar2021ManipulatingTB} consider data poisoning to be of limited harm and gradient-based approach to be too computationally intensive, clean-label attacks based on gradient matching have shown to be both stealth and effective when performing a targeted integrity attack \cite{geiping2021witches}.
To the best of our knowledge, our work is the first to demonstrate a complete availability attack on a neural network using data poisoning.

\paragraph{Availability attacks} have been demonstrated in a range of settings.
As shown in Figure \ref{fig:this-work}, we can compare these settings in term of attacker knowledge (from black box to omniscient) and degree of freedom (from a single constrained interaction to an omnipotent attacker).
The data poisoning literature on neural networks only allows an attacker to craft a poison once to insert it in the training set, with various levels of knowledge.
\cite{geiping2021witches} operate both in \textit{black-box} setting and with sole \textit{model architecture} knowledge.
\cite{munozgonzalez2019poisoning} assume an omniscient attacker and \cite{munozgonzalez2017poisoning} add scenarios with attackers not having access to the model weights or to the training data.
\cite{farhadkhani2022equivalence} assume an attacker who can interfere at the embedding level of the last layer of a neural network.
This particular case is equivalent to attacking a logistic regression which is a convex setting for which an equivalence between data poisoning and gradient attacks holds.
Previous works in data poisoning on neural networks were only able to slightly decrease the performance of the attacked algorithm and have yet to demonstrate a complete availability attack that makes a model truly useless \cite{zhao_clpa_2022, Lu2022IndiscriminateDP}.
Gradient attacks, on the other hand, have established effective ways of making a model utterly useless.
\cite{blanchard2017byzantinetolerant} show how an omniscient attacker sending unconstrained gradients at each iteration can arbitrarily change the model's weights.
\cite{baruch2019little} assume an attacker with access to model's weights and a fraction of the training set (similar to our \textit{auxiliary dataset}).
\cite{elmhamdi2020distributed} suppose an omniscient attacker or one that does not know the legitimate gradients.
In this work, we extend the domain of known availability attacks and demonstrate how they are possible in settings were the attacker knows the model weights and can craft a data poisoning at each iteration in a constrained set or not, similarly to Algorithm 1 in \citet{steinhardt2017certified} (Result A and B in subsection \ref{subsec:results}).

\paragraph{Defenses} against data poisoning have been studied through different approaches such as data sanitization \cite{steinhardt2017certified}, data augmentation \cite{borgnia_strong_2020}, bagging \cite{wang_improved_2022} or pruning and fine-tuning \cite{liu2018finepruning}.
However, the effectiveness of such defenses rely on strong assumptions such as the learner having access to a clean dataset, on the convexity of the loss w.r.t. the model's parameters or the learner's ability to train a very large number of models.
And still, attackers could find ways to break these defenses \cite{koh_stronger_2021}.
Even if a theoretically sound and impenetrable defense mechanism against data poisoning might be impossible \cite{elmhamdi2023impossible,hardt2023algorithmic}, making the attacker's job harder by adding several imperfect yet constraining defenses in a ``Swiss cheese''\footnote{\url{https://en.wikipedia.org/wiki/Swiss_cheese_model}}-like model (as in cybersecurity) is still necessary.

\paragraph{Inverting gradients} has recently been studied in privacy attacks to reconstruct training samples based on the resulting gradient \cite{geiping2020inverting, zhao2020idlg}.
Their central recovery mechanism relies on maximizing a similarity measure $Sim$ between a targeted gradient $g^{(target)}$ that has been computed on the data which the attacker wants to reconstruct and the gradients $G^{(private)} = \{ \nabla_{\theta_{t}} L(h_{\theta_{t}}(x), y) \}_{(x, y) \in S}$ computed on the reconstructed data $S$ and aggregated through the $\textsc{Agg}$ function:
\begin{equation}
    \label{eq:inv-grad-priv-atk}
    S^{priv} \in \arg \max_{S \in (\mathcal{X} \times \mathcal{Y})^{n_{priv}}} Sim(g^{(target)}, \textsc{Agg}( G^{(private)} ) )
\end{equation}
Similarly, we try to recover data points that induce the closest gradient possible to a malicious gradient.
The existence of a solution for (\ref{eq:inv-grad-priv-atk}) in the privacy attack setting is known, since actual data points have given rise to the targeted gradient $g^{(target)}$.
On the contrary, gradient attacks are not actually calculated from a data point hence there is no guarantee that the inversion can find an existing solution.

%% file: sections/03-framework.tex
\section{Framework}
 
\subsection{Learning Setting}

We consider a classification setting where the model is trained on a dataset $D_{train} = \{ (x_{i}, y_{i}) \}_{i=1}^{n}$ sampled from a distribution $\mathcal{D}$ over $\mathcal{X} \times \mathcal{Y}$.
The learner trains a neural network $h_{\theta}$ parametrized by $\theta \in \mathbb{R}^{d}$ with an iterative optimization algorithm on the (non-convex) loss function $L$.
Its goal is to achieve the lowest test loss on a heldout test set $D_{test}$ that is not necessarily sampled from the same distribution $\mathcal{D}$ as the training set.
We formulate the objective as the following optimization problem:
\begin{align*}
    \arg \min_{\theta \in \Theta} \frac{1}{n_{test}} \sum_{(x,y) \in D_{test}} L(h_{\theta}(x), y)
\end{align*}

We consider that learning occurs through a set of $n_b$ \textit{Gradient Generation Units} $\{V_{i}\}_{i=1}^{n_{b}}$ each of which reports a \textit{message} in a set $S^{b}_{t} = \textsc{Message}(D_{train}, t) = \{v_{i, t}\}_{i=1}^{n_{b}}$ at each iteration $t$.
Messages are then aggregated through an aggregator $\textsc{Agg}$ and the model weights are updated using the $\textsc{Update}$ algorithm:
$$
    \theta_{t+1} = \textsc{Update}(\theta_{t}, \textsc{Agg}, S^{b}_{t})
$$ 

This abstraction allows us to represent a large spectrum of learning settings, from centralized learning to fully distributed learning and settings in between (such as federated learning).
\begin{itemize}
    \item In the common centralized setting of training a neural network, the $\textsc{Message}$ operator returns a batch of data points sampled from the training set $S^{b}_{t} = \{ (x_{i, t}, y_{i, t}) \}_{i=1}^{n_{b}}$, the aggregator $\textsc{Agg}$ is the $\textsc{Average}$ of their gradients, computed by the learner and the update algorithm is $\textsc{SGD}$ or $\textsc{Adam}$.
    \item In the federated learning setting, the $\textsc{Message}$ operator returns a batch of gradients (or equivalently model updates) that each worker computed separately, the default aggregator is the $\textsc{Average}$ of the messages and the update algorithm is $\textsc{FederatedAveraging}$ \cite{mcmahan2017communication} or $\textsc{LocalSGD}$ \cite{stich2018local}.
\end{itemize}
The choice for the $\textsc{Message}$ operator boils down to entrusting the calculation of gradients to the workers or not.
This, in turn, gives different degrees of freedom for a malicious \textit{Gradient Generation Unit} to influence the training: in the scope of this paper, respectively gradient attacks or data poisoning\footnote{The notion of \textit{Gradient Generation Units} can be used as an abstraction for any source of input that is itself a gradient, a model update, or that can be later transformed to a gradient, such as a training data point, a machine in distributed learning or a user account in a social media platform, depending on the level of granularity that is considered.}.

In our experiments, we consider an image classification task on the CIFAR10 dataset on which the attacker can tamper with messages (data point or gradient depending on the learning setting). We use the $\textsc{SGD}$ and $\textsc{Adam}$ update algorithms as well as $\textsc{Average}$ and $\textsc{MultiKrum}$  \cite{blanchard2017byzantinetolerant} aggregators. $\textsc{MultiKrum}_{f<0.5}$ is a robust aggregator which can withstand a ratio of malicious elements up to $f$ and is defined for a set of vectors $\{v_{i}\}_{i=1}^{n}$ as the average of the $n(1-f)-2$ vectors minimizing the score function $s(i)=\sum_{i \rightarrow j} \| v_{i} - v_{j} \|^{2}$, with $i \rightarrow j$ the indices of the $n(1 - f) - 2$ closest vectors to $v_{i}$.

The choices of update and aggregator we make are important since in the Robust Distributed Learning literature, the update algorithm and aggregator are both considered defense mechanisms \cite{elmhamdi2020distributed}.

\subsection{Threat model}

\begin{figure}[t]
    \centering
    \includegraphics[width=0.7\linewidth]{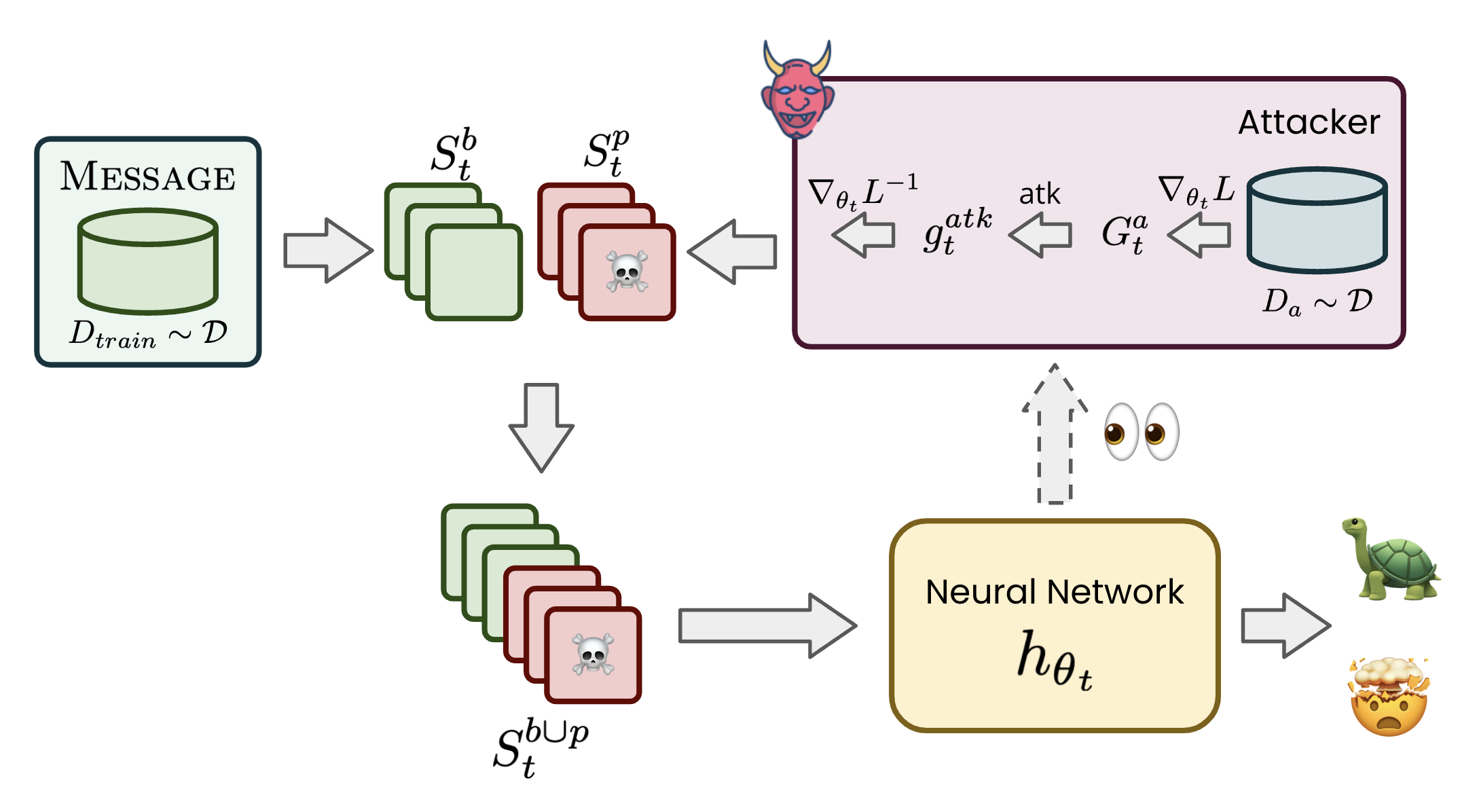}
    \caption{Threat model. The attacker has access to $\theta$ but does not have access to the batch $S^{b}_{t}$ and uses an auxiliary dataset $D_{a}$ to craft $S^{p}_{t}$ the set of poisoned messages. Both the batch and the poisons set are gathered into $S^{b \cup p}_{t}$. The attacker's goal is either to slow down the training or attack the model's availability.}
    \label{fig:threat-model-dp}
\end{figure}

We consider the following threat model, in which the attacker:
\begin{itemize}[itemsep=0.05em]
    \item has knowledge of the weights of the model $\theta_{t}$, the update algorithm $\textsc{Update}$, the aggregator $\textsc{Agg}$ function and the message operator $\textsc{Message}$ as in among others, \cite{farhadkhani2022equivalence,blanchard2017byzantinetolerant,yin2021byzantinerobust,shafahi2018poison}; 
    \item does not have access to the batch $S^{b}_{t}$, unlike the stronger standard assumption of omniscient attacker in the robust distributed learning literature \cite{blanchard2017byzantinetolerant, elmhamdi2020distributed, lilisu2018distribstatml, yin2021byzantinerobust};
    \item has access to an auxiliary dataset $D_{a} \sim \mathcal{D}$ that is a surrogate to the unobservable training set. This is a standard assumption when not assuming omniscient attackers, as in \cite{mhamdi2018hidden,farhadkhani2022equivalence, elmhamdi2020distributed}; 
    \item has the control over a ratio $\alpha$ of \textit{Gradient Generation Units} and the ability to append a set of arbitrarily crafted poisoned messages (gradients or data points) $S^{p}_{t} = \{v^{p}_{i, t}\}_{i=1}^{n_{p}}$ to the clean batch $S^{b}_{t}$ at each iteration $t$ up to a level $\alpha$ of contamination (i.e. ${ | S^{p}_{t}  |}/{ | S^{b \cup p}_{t}  |}=\alpha$), as in \cite{blanchard2017byzantinetolerant,mhamdi2018hidden,yin2021byzantinerobust} for gradients attacks, in \cite{steinhardt2017certified} for data poisoning or in \cite{farhadkhani2022equivalence} for both gradients attacks and data poisoning; 
    \item constrains itself\footnote{Result A in subsection~\ref{subsec:results} is our strongest result and is obtained under this constraint on the attacker. Result B is~\ref{subsec:results} is obtained without this constraint and serves to further understand the expressivity of data poisoning from inverted gradient attacks.} to only crafting poisons that are in a feasible domain $S^{p}_{t} \in \mathcal{F}^{n_{p}}$, depending on the task and data structure. For instance, for an image classification task of $H \times W$ RGB pixels encoded on 3 bytes and labels are between 1 and $C$, we set $\mathcal{F}$ to be the set of possible such sized images and labels: $\mathcal{F}=[0..255]^{H \times W \times 3} \times [1..C]$ as in \cite{farhadkhani2022equivalence}, but without restricting ourselves to convex settings.
\end{itemize}

The attacker's goal is to perform an \textbf{availability attack:} degrading the learner's performances as much as possible.
We also consider \textbf{slow down attacks}, in which the attacker tries to stall the learning procedure as much as possible.
This attack is important, considering the costs for training large models and the direct financial repercussion for slowing them down.

This threat model is inspired from gradient attacks and assumes much more communication between the attacker and the model than traditional data poisoning threat models.
The latter leads to poor availability attacks performance.
We hypothesized that once we allowed for the same amount of interactions between gradient attacks and data poisoning, the two methods would display far less discrepancy of destructive power than what is currently thought. 

%% file: sections/04-method.tex
\section{Method}

At each iteration $t$, the attacker, controlling $p$ \textit{Gradient Generation Units}, computes an auxiliary batch $S^{a}_{t} = \textsc{Message}(D_{a}, t)$ and constructs $S^{p}_{t}$ such that $\textsc{Update}(\theta_{t}, \textsc{Agg}, S^{a}_{t} \cup S^{p}_{t})$ gives poorer performance on $D_{a}$ than $\textsc{Update}(\theta_{t}, \textsc{Agg}, S^{a}_{t})$.
This should, in turn, also decrease the model's performances on $D_{train}$ under the assumption it follows the same distribution as $D^{a}$.

\subsection{Gradient attacks}

When the $\textsc{Message}$ operator outputs gradients, %
a malicious \textit{Gradient Generation Units} can participate in $S^{b}_{t}$ with a gradient attack in $\mathbb{R}^{d}$.
At each iteration $t$, the attacker computes the gradients over the auxiliary dataset $D_{a}$ since they cannot access the batch $S^{b}_{t}$.
Let $S^{a}_{t} = \{ g^{a}_{t,i} \}_{i=1}^{n_{a}} = \{ \nabla_{\theta_{t}}L(h_{\theta_{t}}(x), y) \}_{(x,y) \in D_{a}}$ be the set of per-sample auxiliary gradients and $g^{a}_{t} = \frac{1}{n_{a}} \sum_{g \in S^{a}_{t}} g$ the averaged auxiliary gradient.
Similarly, $g^{b}_{t} = \frac{1}{n_{b}} \sum_{g \in S^{b}_{t}} g$ denotes the averaged clean batch gradient.
The attacker constructs a set of $n_{p}$ attacking gradients $G^{atk}_{t}$ based on $S^{a}_{t}$ and sends them as message $S^{p}_{t}$.
We denote $S^{b \cup p}_{t} = S^{b}_{t} \cup S^{p}_{t}$ the poisoned batch of gradients, $S^{a \cup p}_{t} = S^{a}_{t} \cup S^{p}_{t}$ the poisoned auxiliary gradients and $g^{b \cup p}_{t}$ and $g^{a \cup p}_{t}$ their respective averages.
We consider several gradient attacks to perform an availability attack.

\begin{itemize}
    \item \textbf{Gradient Ascent} (GA) \cite{blanchard2017byzantinetolerant}: the attacker sends a gradient such that the averaged poisoned gradient is anti-collinear with the mean clean gradients. This provokes a gradient ascent step with the $\textsc{SGD}$ update rule, $\textsc{Average}$, and $\lambda>0$:%
    \begin{align*}
        \mathbb{E}_{S^{b}_{t}}[\theta_{t+1}]    &= \theta_{t} - \eta \mathbb{E}_{S^{b}_{t}}[g^{b \cup p}_{t}] \\
                                                &= \theta_{t} + \eta \lambda \mathbb{E}_{S^{b}_{t}}[g^{b}_{t}]
    \end{align*}
    \item \textbf{Orthogonal Gradient} (OG): the attacker sends a gradient such that the averaged poisoned gradient is orthogonal to the mean clean gradient. This should cause the training to stall.
    \item \textbf{Little is Enough} (LIE) \cite{baruch2019little, Shejwalkar2021ManipulatingTB}: the attacker sends the mean clean gradient deviated by a strategically chosen amount times the coordinate-wise standard deviation of the clean gradients, with $\sigma[j] = \sqrt{Var(\{ g[j] \}_{g\in S^{a}_{t}})}$.
    \begin{align}\label{eq:lie}
        && g^{atk}_{t} &= g^{a}_{t} - z^{max} \sigma \\
       \text{where} && z^{max} &\in \arg \max_{z \in \mathbb{R}^{*}_{+}} \left \| A^{a \cup p}_{t} - A^{a}_{t} \right \|, \notag \\
       \text{and} && A^{a \cup p}_{t} &= \textsc{Agg}( S^{a}_{t} \cup \{g^{a}_{t} - z \sigma\}_{i=1}^{n_{p}} ) \notag \\
       && A^{a}_{t} &= \textsc{Agg}( S^{a}_{t} ) \notag
    \end{align}
    This attack has been shown to be effective against $\textsc{MultiKrum}$ aggregator.
    Note that contrary to \cite{baruch2019little} and similarly to \cite{Shejwalkar2021ManipulatingTB}, we use an adaptive approach and choose $z^{max}$ to maximize the divergence of the poisoned gradients on the aggregation $A^{a \cup p}_{t}$.
\end{itemize}

\subsection{Data poisoning}
\label{subsec:data_poisoning_method}

When the $\textsc{Message}$ operator outputs data samples, malicious \textit{Gradient Generation Units} are expected to participate via similarly structured messages, i.e. data poisoning.
In our experiments, on CIFAR10 and its $32 \times 32$ RGB images, it leaves room for an attacker to participate in the training process via messages crafted in $[0, 255]^{32 \times 32 \times 3}$.
This gives only $32 \times 32 \times 3 = 3,072$ degrees of freedom, far less than commonly used deep learning models' number of parameter, hinting that data poisoning should be less expressive than gradient attacks.
Models' non-linearities further constrain the dimension of the image space of the gradient operator (as illustrated in Figure~\ref{fig:image-grad}).
Data poisoning is allegedly harder as it constitutes a far more constrained problem than gradient attacks (in which the attacker can send an arbitrary gradient from $\mathbb{R}^{d}$).

At each iteration $t$, the attacker computes a given gradient attack $g^{atk}_{t}$ in the same manner as above and inverts the gradient operator to compute an associated set of data points $S^{p}_{t} = \{(x^{p}_{i, t}, y^{p}_{i, t})\}_{i=1}^{n_p}$ such that $\frac{1}{n_p} \sum_{i=1}^{n_{p}} \nabla_{\theta_{t}} L(h_{\theta_{t}}(x^{p}_{i, t}), y^{p}_{i, t}) = g^{atk}_{t}$.
That set of data points is then sent a message $S^{p}_{t}$ and appended to $S^{b}_{t}$, like in the gradient attack case.\\
Similarly to what is done in privacy attacks, we optimize a similarity measure between the gradient attack and the gradients calculated on $S^{p}_{t}$.
Contrary to privacy attacks where the existence of a solution is known\footnote{In privacy attacks, the gradient being inverted was produced by the data point that the attacker is trying to retrieve.}, there is no guarantee that the inversion can find an existing solution, let alone an effective data poison.

We can formulate our data poisoning as an optimization problem where the attacker aims to minimize a poisoning function $f_{p}$. It characterizes the dissimilarity of the data poison gradients $G^{p}_{t} = \{ \nabla_{\theta_{t}} L(h_{\theta_{t}}(x), y) \}_{(x, y) \in S^{p}_{t}}$ with a gradient attack $g^{atk}_{t}$ calculated from the auxiliary dataset gradients $G^{a}_{t} = \{ \nabla_{\theta_{t}} L(h_{\theta_{t}}(x), y) \}_{(x, y) \in S^{a}_{t}}$ (since the attacker cannot have access to the batch $S^{b}_{t}$) over a feasible domain $\mathcal{F}$:
\begin{equation*} \label{eq:dp-general}
    S^{p}_{t} \in \arg \min_{S \in \mathcal{F}^{n_{p}}} f_{p}(h_{\theta_{t}}, G^{a}_{t}, S)
\end{equation*}
Since characterizing the image space of the gradient operator on the loss function $L$ is a difficult task, we cannot know beforehand if a vector can have an antecedent data point through the gradient operator.
Together with constraints on feasibility, finding data poisons which exactly reproduce the gradient attack might be impossible.
Since $\mathcal{I}_{\nabla \mathcal{F}} = \nabla_{\theta_{t}} L(h_{\theta_{t}}(\mathcal{F}_{\mathcal{X}}), \mathcal{F}_{\mathcal{Y}}) \subseteq \nabla_{\theta_{t}} L(h_{\theta_{t}}(\mathcal{X}), \mathcal{Y}) \subseteq \mathbb{R}^{d}$, the set $\mathcal{I}_{\nabla \mathcal{F}}$  might not cover all the possible candidates for an effective gradient attack, if not any (Figure \ref{fig:image-grad}).
Thus, achieving an effective data poisoning that performs similarly to a gradient attack is non trivial.

We show in our experiments that an attacker can still produce data poisons which have significant impact on the training procedure.
Poisons are iteratively updated to minimize the poisoning objective $f_{p}$ using the \texttt{Adam} optimizer and are projected on the feasible set $\mathcal{F}$ at each iteration of the poisoning optimization by clipping.
Table \ref{tab:grad-atk-dp-atk} details the formulas used to determine the gradient attacks and their equivalent poisoning functions in the data poisoning case.

\renewcommand{\arraystretch}{1.5}
\begin{table*}[ht]
    \caption{Gradient attacks formula and their equivalent data poisoning objective functions to be optimized. $g^{p}, g^{a}, g^{a \cup p}$ are respectively the averaged gradients computed on the data poisons, on the auxiliary dataset, and the weighted average between them. $\cos$ is the cosine similarity. $z^{max}$ and $\sigma$ are defined as in eq. \ref{eq:lie}.}
    \label{tab:grad-atk-dp-atk}
    \centering
    \begin{tabular}[]{|c|c||c|}
        \hline
          &  \multirow[c]{2}{*}{\makecell{\textbf{Gradient attack}\\$g^{p} \in \mathbb{R}^{d}$ s.t.}} & \multirow[c]{2}{*}{\makecell{\textbf{Our data poisoning attack}\\$f_{p}$}} \\
         & & \\
        \hline
        Gradient Ascent & $\cos(g^{a \cup p}, g^{a}) = -1$ & $\cos\left ( g^{a \cup p}, g^{a} \right )$ \\
        \hline
        Orthogonal Gradient & $\langle g^{a \cup p}, g^{a} \rangle = 0$ & $\left \| \cos\left ( g^{a \cup p}, g^{a} \right ) \right \|^{2}$ \\
        \hline
        Little is Enough & $g^{p} = g^{a} - z^{max} \times \sigma$ & $\left \| g^{p} - g^{a} + z^{max} \sigma \right \|^{2}$ \\
        \hline
    \end{tabular}
\end{table*}
\renewcommand{\arraystretch}{1}

%% file: sections/05-experiments.tex
\section{Experiments}

\subsection{Preliminary experiment}

\begin{figure}[h]
    \centering
    \hfill
    \vskip 0pt
    \begin{subfigure}[b]{0.3\linewidth}
        \includegraphics[width=\linewidth]{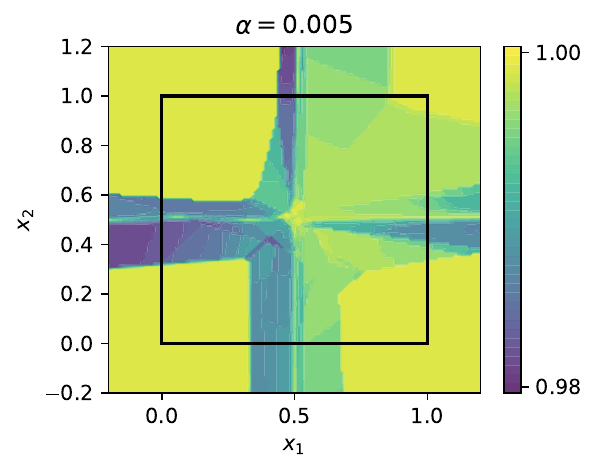}
    \end{subfigure}
    \hfill
    \begin{subfigure}[b]{0.30\linewidth}
        \includegraphics[width=\linewidth]{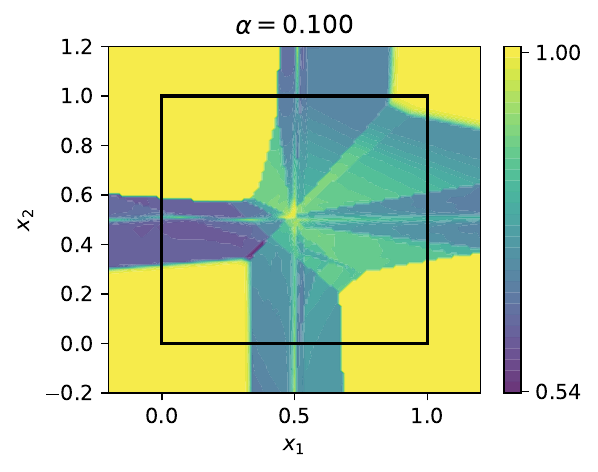}
    \end{subfigure}
    \hfill
    \begin{subfigure}[t]{0.25\linewidth}
        \vspace{-4cm}
        \includegraphics[width=\linewidth]{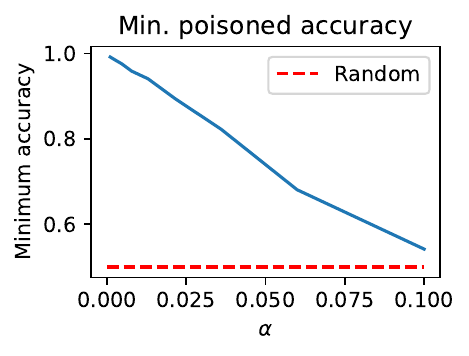}
    \end{subfigure}
    \caption{Landscape of accuracies after 1 poisoned step with the respective poison with two levels of contamination $\alpha= 0.005$ (\textbf{left)} and $\alpha=0.1$ (\textbf{center}). The black squares represent the border of the feasible domain $\mathcal{F}$. \textbf{Right:} The minimum accuracy in the landscape for different levels of contamination $\alpha$.}
    \label{fig:acc_xor_pois}
\end{figure}

To give a better intuition of the capabilities of data poisoning, we present a preliminary experiment on the \texttt{XOR} operator classification task.
Sampling a point $x$ in $[0, 1]^{2}$, its label is $y = \mathbbm{1}\{x[0] > 0.5\} \oplus \mathbf{1}\{x[1] > 0.5\}$ with $\mathbbm{1}$ the indicator function and $\oplus$ the \texttt{XOR} operator.
We generate a set of possible poisons by regularly sampling on the $[0, 1]^{2}$ grid and labeling them with \texttt{XOR} then \textbf{flipping} their labels.
After training a multilayer perceptron with Gradient Descent (i.e. full batch) on 1000 data points, we perform a single step of gradient descent on the data poisoned with one of the possible poisons, repeated as to reach a contamination level $\alpha$.
The accuracies obtained for all poisons are illustrated in Figure~\ref{fig:acc_xor_pois}.

A 10\% level of contamination is already enough to bring the model close to the level of a random guess, which suggests that a single step on poisoned data is enough to obtain an availability attack on neural network for a task such \texttt{XOR}. This proof of concept on \texttt{XOR} could however be argued to benefit from the inner volatility of binary classification based on binary inputs, where merely flipping labels can translate into inflated consequences in the model space. In the next subsection, we present our main empirical results on more complex differentiable tasks of image classification, testing the realism of our correspondence between prominent gradient attacks and data poisoning, i.e. testing our proposed gradient attacks inversion in Table~\ref{tab:grad-atk-dp-atk}).
\subsection{Experimental setup}

\paragraph{Model \& dataset} We demonstrate our poisoning procedure on a custom convolutional neural network (described in Table \ref{tab:cnn-arch} in the Appendix) and on Vision Transformers models (\texttt{ViT-tiny} models with patch size 8) trained for 50 epochs on the CIFAR10 dataset partitioned in training, validation, and auxiliary datasets.
We use different optimization algorithms and aggregation rules to train the models: $\textsc{SGD}$ \& $\textsc{Average}$, $\textsc{Adam}$ \& $\textsc{Average}$, $\textsc{SGD}$ \& $\textsc{MultiKrum}$ (with different levels of data truncation $f \in \{ 0.1, 0.2, 0.4 \}$).
Since $\textsc{MultiKrum}$ is used as a defense mechanism, we expect it to be robust to the attacks that operate within its working assumptions.
However, the Little is Enough attack has specifically been designed to circumvent these assumptions when using gradient attacks.
We thus expect Little is Enough to bypass $\textsc{MultiKrum}$ in the gradient attacks situation.

\paragraph{Baseline} In every setting, the learning rate is fixed to the value were the learner achieves the best performances without any poisoning to set a baseline for the performances of the model. We then measure the decay in performances caused by an attack. Each setting of this experiment is run 4 times for better statistical significance. Each run has a different model and poisons initialization, and dataset split. Full results can be seen in the Appendix.

\paragraph{Attacks} In every setting, we perform either one of the considered attacks either via a gradient attack, or a data poisoning attack as specified in section \ref{subsec:data_poisoning_method}.
The $n_{p}$ crafted poisons are added to the batch of size $n_{b}$ at each iteration so that $\frac{n_{p}}{n_{b}+n_{p}} = \alpha \in [0.01, 0.48]$.
Since the gradients induced by the poisoned data mimic the malicious gradients, we expect our data poisoning is at best as good as the associated gradient attack.
Gradient attacks should thus be a topline for our data poisoning attack.

\subsection{Results}
\label{subsec:results}

\subsubsection{Gradient attacks} Table \ref{tab:grad-avail-best-val} shows that gradient attacks do perform an availability attack, bringing the models' performances down to random-level.
We also notice that the $\textsc{MultiKrum}$ aggregation rule does act as a defense mechanism, for levels of contamination lower than its tolerance parameter $f$.
However, as expected, this defense is well circumvented by the Little is Enough attack.

\begin{table*}[ht]
    \caption{Best validation accuracy under different attacks with different update rules, different aggregation functions and different levels of contamination $\alpha$. A high validation accuracy (colored in apricot) indicates a failed attack. A low validation accuracy (colored in pale green) indicates a successful attack.}
    \label{tab:grad-avail-best-val}
    \centering
    \begin{tabular}{ccccccccc}
    \toprule
     \multirow[c]{2}{*}{\makecell{$\textsc{Update}$;\\$\textsc{Agg}$}} & \multirow[c]{2}{*}{\makecell{Attack}} & \multicolumn{7}{c}{\boldmath$\alpha$\unboldmath} \\
     & &       0.01 &       0.05 &       0.10 &       0.20 &       0.30 &       0.40 &       0.48 \\
    \midrule
    \multirow[c]{3}{*}{\makecell{$\textsc{Adam}$;\\$\textsc{Average}$}} & GA & \cellcolor{Apricot!10!LightAquaGreen}10.0 &  \cellcolor{Apricot!10!LightAquaGreen}9.8 &  \cellcolor{Apricot!10!LightAquaGreen}10.0 &  \cellcolor{Apricot!10!LightAquaGreen}10.0 &  \cellcolor{Apricot!10!LightAquaGreen}10.0 &  \cellcolor{Apricot!10!LightAquaGreen}10.1 &  \cellcolor{Apricot!10!LightAquaGreen}10.0 \\
        & OG &  \cellcolor{Apricot!10!LightAquaGreen}9.9 &  \cellcolor{Apricot!10!LightAquaGreen}9.9 &  \cellcolor{Apricot!10!LightAquaGreen}10.1 &  \cellcolor{Apricot!10!LightAquaGreen}10.1 &  \cellcolor{Apricot!10!LightAquaGreen}9.8 &   \cellcolor{Apricot!10!LightAquaGreen}10.0 & \cellcolor{Apricot!10!LightAquaGreen}10.1 \\
        & LIE &  \cellcolor{Apricot!10!LightAquaGreen}10.2 &  \cellcolor{Apricot!10!LightAquaGreen}10.3 &  \cellcolor{Apricot!10!LightAquaGreen}10.1 &   \cellcolor{Apricot!10!LightAquaGreen}10.2 &  \cellcolor{Apricot!10!LightAquaGreen}10.1 &   \cellcolor{Apricot!10!LightAquaGreen}10.1 &  \cellcolor{Apricot!10!LightAquaGreen}10.0 \\
    \hline
    \multirow[c]{3}{*}{\makecell{$\textsc{SGD}$;\\$\textsc{Average}$}} & GA &  \cellcolor{Apricot!10!LightAquaGreen}10.1 &  \cellcolor{Apricot!10!LightAquaGreen}10.1 &  \cellcolor{Apricot!10!LightAquaGreen}10.1 &   \cellcolor{Apricot!9!LightAquaGreen}9.9 &  \cellcolor{Apricot!10!LightAquaGreen}10.0 &  \cellcolor{Apricot!10!LightAquaGreen}10.0 &  \cellcolor{Apricot!10!LightAquaGreen}10.0 \\
        & OG &  \cellcolor{Apricot!10!LightAquaGreen}10.2 &  \cellcolor{Apricot!10!LightAquaGreen}10.2 &   \cellcolor{Apricot!10!LightAquaGreen}10.0 &   \cellcolor{Apricot!9!LightAquaGreen}9.9 &  \cellcolor{Apricot!10!LightAquaGreen}10.1 &  \cellcolor{Apricot!10!LightAquaGreen}10.1 &  \cellcolor{Apricot!10!LightAquaGreen}10.2 \\
        & LIE &  \cellcolor{Apricot!10!LightAquaGreen}10.2 &   \cellcolor{Apricot!9!LightAquaGreen}9.9 &  \cellcolor{Apricot!10!LightAquaGreen}10.2 &   \cellcolor{Apricot!9!LightAquaGreen}9.9 &   \cellcolor{Apricot!9!LightAquaGreen}9.9 &   \cellcolor{Apricot!10!LightAquaGreen}10.0 &   \cellcolor{Apricot!10!LightAquaGreen}10.0 \\
    \hline
    \multirow[c]{3}{*}{\makecell{$\textsc{SGD}$;\\$\textsc{MultiKrum}_{f=0.1}$}} & GA & \cellcolor{Apricot!65!LightAquaGreen}65.1 &  \cellcolor{Apricot!63!LightAquaGreen}63.2 &  \cellcolor{Apricot!33!LightAquaGreen}33.7 &  \cellcolor{Apricot!10!LightAquaGreen}10.1 &  \cellcolor{Apricot!10!LightAquaGreen}10.1 &  \cellcolor{Apricot!10!LightAquaGreen}10.2 &  \cellcolor{Apricot!10!LightAquaGreen}10.2 \\
        & OG &  \cellcolor{Apricot!65!LightAquaGreen}65.1 &  \cellcolor{Apricot!65!LightAquaGreen}65.3 &  \cellcolor{Apricot!65!LightAquaGreen}65.9 &  \cellcolor{Apricot!10!LightAquaGreen}10.0 &  \cellcolor{Apricot!10!LightAquaGreen}10.1 &  \cellcolor{Apricot!10!LightAquaGreen}10.2 &   \cellcolor{Apricot!9!LightAquaGreen}9.9 \\
        & LIE &  \cellcolor{Apricot!41!LightAquaGreen}41.7 &  \cellcolor{Apricot!15!LightAquaGreen}15.0 &  \cellcolor{Apricot!10!LightAquaGreen}10.4 &   \cellcolor{Apricot!9!LightAquaGreen}9.9 &  \cellcolor{Apricot!10!LightAquaGreen}10.0 &   \cellcolor{Apricot!9!LightAquaGreen}9.9 &  \cellcolor{Apricot!10!LightAquaGreen}10.1 \\
    \bottomrule
    \end{tabular}
\end{table*}

\subsubsection{Data poisoning}
\begin{figure*}[th]
    \centering
    \hfill
    \begin{subfigure}[b]{0.45\linewidth}
        \centering
        \includegraphics[width=0.55\linewidth]{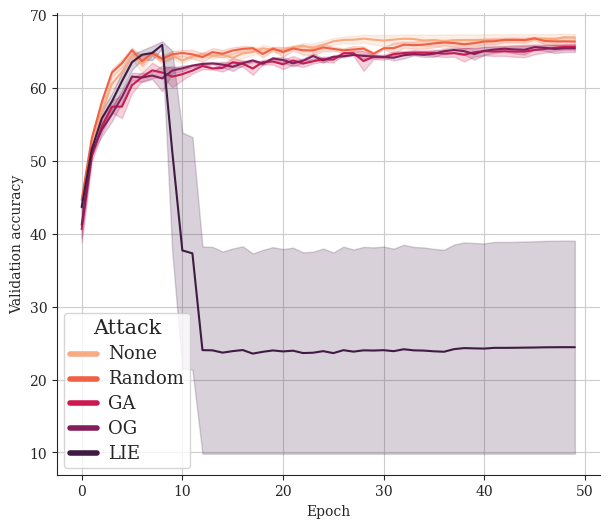}
        \caption{Comparison of different attacks\\ at $\alpha = 0.01$.}
        \label{fig:dp-avail-atk-val-acc-a001}
    \end{subfigure}
    \hfill
    \begin{subfigure}[b]{0.46\linewidth}
        \centering
        \includegraphics[width=\linewidth]{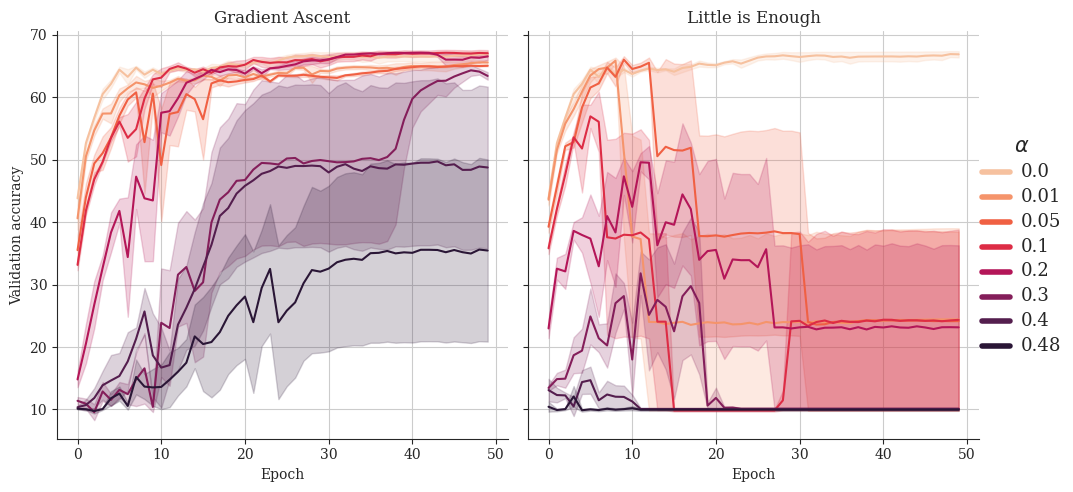}
        \caption{Comparison of different levels of contamination for the Gradient Ascent \& Little is Enough attacks.}
        \label{fig:dp-avail-atk-val-acc-cosim-little}
    \end{subfigure}
    \hfill
    \caption{Validation accuracies during training in the $\textsc{SGD}$ \& $\textsc{Average}$ setting under different attacks and different level $\alpha$ of contamination. Error bars represent the standard error.}
\end{figure*}

After performing the equivalent data poisoning attacks, the observed effects range from a slowdown of the training procedure to its complete halt and up to degrading the performances down to random levels.
Figure \ref{fig:dp-avail-atk-val-acc-a001} compares the different attacks with a random data poisoning (uniformly sampled in $[0,1]$) at $\alpha = 0.01$ with its baseline counterpart.
While the Gradient Ascent and Orthogonal attacks behave similarly and at most only slightly slow down the training at most, Little is Enough attack strongly degrades performances, even at such low levels of contamination.
Figure \ref{fig:dp-avail-atk-val-acc-cosim-little} shows for the Gradient Ascent and Little is Enough attacks that the higher the level of contamination, the stronger the effect: validation accuracy increases slower or plummets earlier.
Therefore, our main result is the following.

\begin{figure}[h]
    \centering
    \includegraphics[width=\linewidth]{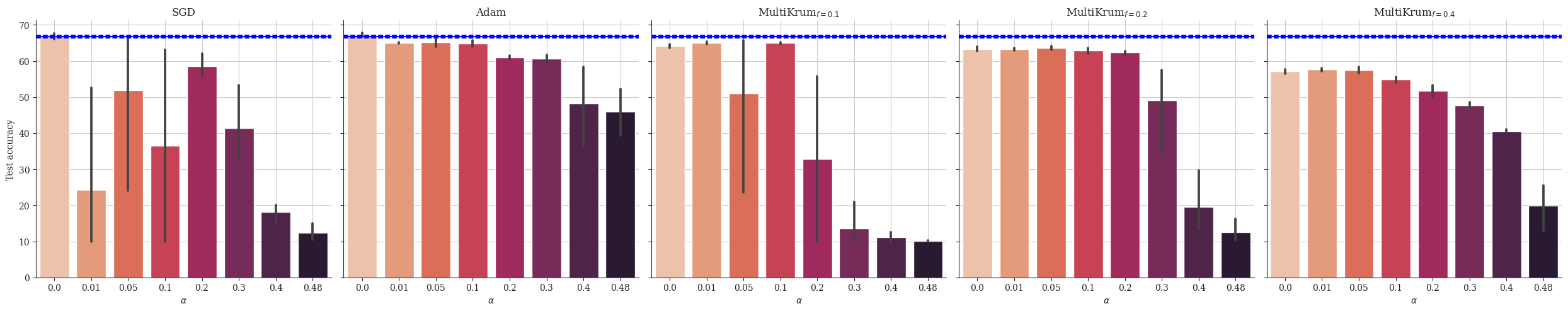}
    \caption{Test accuracy of the CNN model which achieved the best validation accuracy under the Little is Enough attack. Each column represents a different setting of update function and aggregation rule. The blue line is the test accuracy obtained with ($\textsc{SGD}$, $\textsc{Average}$) setting and no attack. Error bars are the standard errors.}
    \label{fig:dp-avail-best-test-acc-little}
\end{figure}

\paragraph{Availability attacks (Result A).} To fairly compare each setting under that attack, each model has been evaluated on the test set with the weights achieving the best validation accuracy.
Figure \ref{fig:dp-avail-best-test-acc-little} shows that for the $\textsc{SGD}$ \& $\textsc{Average}$ learner under the Little is Enough attack, \boldmath$\alpha = 0.01$\unboldmath \textbf{ is enough to significantly degrade the model's performance}.
Among the 4 runs, 3 ended up diverging while every poison was \textbf{in the feasible set} $\mathcal{F}$. Figure \ref{fig:dp-avail-sgd-little1e0-a001-50pois} in the Appendix shows 50 of the first 500 poisons crafted by the attacker in one of the diverging run.
Similarly, in the $\textsc{SGD}$ \& $\textsc{MultiKrum}_{f=0.1}$ setting, the Little is Enough attack with contamination level \boldmath$\alpha = 0.05$\unboldmath \textbf{ finds data poisons that circumvent the robust aggregation rule} and drastically reduce the model's performance.

\paragraph{Comparison of update rules.} Figure \ref{fig:dp-avail-best-test-acc-little} shows that our data poisoning procedure failed at performing an availability attack against the $\textsc{Adam}$ update rule.
Because $\textsc{Adam}$ normalizes the aggregated gradients, the training cannot diverge as abruptly as with $\textsc{SGD}$.
However, the slow down attack can still be observed, meaning that the data poisoning procedure could find a solution that perturbs the total gradient enough to slow down convergence, but not enough to completely halt it.
Table \ref{tab:dp-tr-time-atk-val-acc-adam} shows that higher levels of contamination lead to a lower best validation accuracy but most importantly to a higher number of epochs (hence a longer time) to reach it.

\renewcommand{\arraystretch}{1.5}
\begin{table}[t]
    \centering
    \caption{Epoch at which the best validation accuracy is reached (validation accuracies in parenthesis) for the CNN model with $\textsc{Adam}$ optimizer and $\textsc{Average}$ under different data poisoning attacks. Higher levels of contamination induce slower training and lower performances.}
    \begin{tabular}{llll}
        \toprule
        \boldmath$\alpha$\unboldmath & \textbf{GA} & \textbf{OG} & \textbf{LIE} \\
        \midrule
        \textbf{0}     & ~4 (67.1)   & ~4 (67.1) & ~4 (67.1) \\
        \textbf{0.01}     & ~5 (66.4)   & ~5 (66.4) & ~6 (66.3) \\
        \textbf{0.05}     & ~7 (66.5)   & ~7 (65.3) & ~7 (66.3) \\
        \textbf{0.1}     & 19 (63.6)   & 14 (63.9) & 10 (65.7) \\
        \textbf{0.2}     & 34 (61.2)   & 17 (62.5) & 22 (62.1) \\
        \bottomrule
    \end{tabular}
    \label{tab:dp-tr-time-atk-val-acc-adam}
\end{table}
\renewcommand{\arraystretch}{1}

\paragraph{Data poisoning against a robust aggregation rule.} Table \ref{tab:dp-avail-sel-pois-ratio-mk01} shows that with the $\textsc{MultiKrum}_{f=0.1}$ aggregation rule (which filter gradients), the attacker still manages to have some of its messages not filtered out.
For $\alpha > f$, the aggregation rule does not play a defensive role anymore as per its functioning conditions.
On the other hand, for $\alpha$ below this point, the Little is Enough attack displays significantly higher selection rates than the other attacks.
This means that the attacker manages to produce data poisons whose gradients deceive $\textsc{MultiKrum}_{f=0.1}$, similarly to the gradient version of the attack which is particularly designed for this purpose.
However Figure \ref{fig:dp-avail-best-test-acc-little} shows that a higher selection rate does not necessarily mean success of attack.
Even if the attacker sometimes manages to successfully attack the model, $\textsc{MultiKrum}$ overall enhances the robustness of the model while slightly degrading its performances.

\begin{table}
    \caption{Poison selection rates in the $\textsc{SGD}$ \& $\textsc{MultiKrum}_{f=0.1}$ averaged over all runs and all epochs. Each row corresponds to a different attack. Standard deviations in parenthesis.}
    \label{tab:dp-avail-sel-pois-ratio-mk01}
    \centering
    \begin{tabular}{rcccc}
    \toprule
                                     & \multicolumn{4}{c}{\boldmath$\alpha$\unboldmath} \\
                                     & \textbf{0.01} & \textbf{0.05}        & \textbf{0.1}         & \textbf{0.2} \\
    \midrule
    \multirow[c]{2}{*}{\textbf{GA}}  & 0.0           & $2 \times 10^{-4}$   & $2 \times 10^{-3}$   & \textbf{0.91} \\
                                     & (0.0)         & ($4 \times 10^{-4}$) & ($2 \times 10^{-3}$) & (0.08) \\
    \hline
    \multirow[c]{2}{*}{\textbf{OG}}  & 0.0           & $3 \times 10^{-4}$   & 0.18                 & 0.85 \\
                                     & (0.0)         & ($1 \times 10^{-3}$) & (0.1)                & (0.05) \\
    \hline
    \multirow[c]{2}{*}{\textbf{LIE}} &\textbf{0.92}  & \textbf{0.91}        & \textbf{0.97}        & 0.87 \\
                                     &(0.07)         & (0.13)               & (0.02)               & (0.06) \\
    \bottomrule
    \end{tabular}
\end{table}

\begin{figure*}[t]
    \centering
    \begin{subfigure}{0.23\linewidth}
        \centering
        \includegraphics[width=\linewidth]{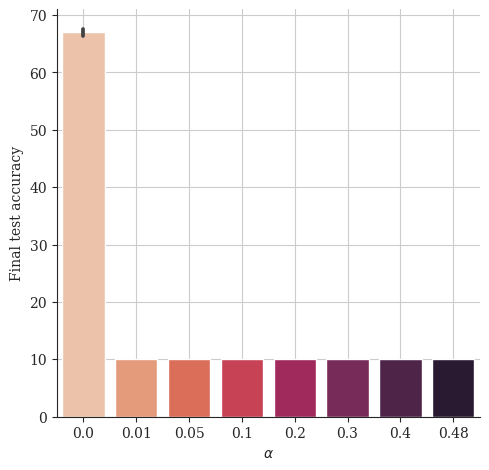}
        \caption{$\mathcal{F}_{\mathcal{X}\text{free}}$}
    \end{subfigure}
    \begin{subfigure}{0.23\linewidth}
        \centering
        \includegraphics[width=\linewidth]{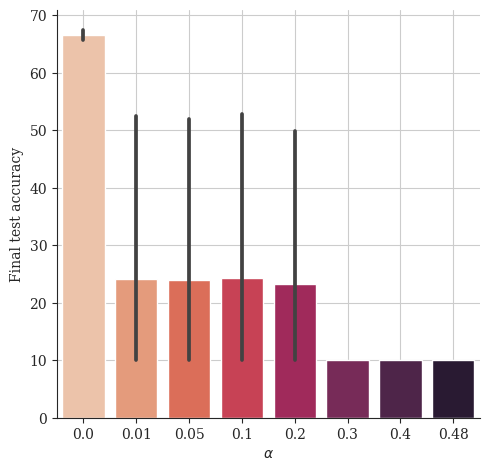}
        \caption{$\mathcal{F}_{\mathcal{X}\text{img}}$}
    \end{subfigure}
    \begin{subfigure}{0.23\linewidth}
        \centering
        \includegraphics[width=\linewidth]{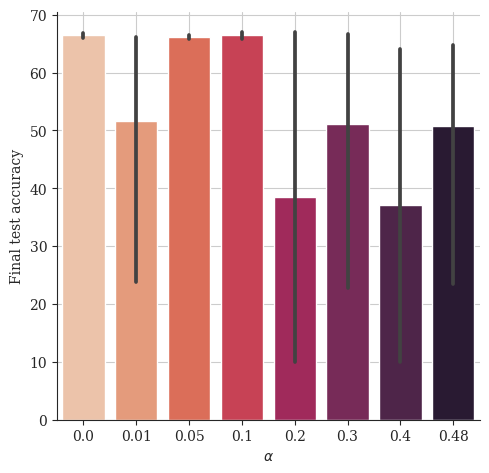}
        \caption{$\mathcal{F}_{\mathcal{X}\text{nei}}$}
    \end{subfigure}
    \hfill
    \caption{Final test accuracies for the $\textsc{SGD};\textsc{Average}$ setting under the Little is Enough attack for different feasible sets. 
    Error bars are the standard errors.}
    \label{fig:dp-avail-atk-test-acc-sgd-feasible-set}
\end{figure*}

\paragraph{Influence of the feasible set (Result B).} As the feasible set $\mathcal{F}_{\mathcal{X}}$ changes, the attacker will only be allowed to converge (in case of convergence) towards different poisons. We compare three increasingly restrictive feasible sets to determine their influence on the success of the attack:
\begin{itemize}[nolistsep]
    \item Constraint-free set: the feasible set is simply the input domain $\mathcal{F}_{\mathcal{X}\text{free}} = \mathbb{R}^{H \times W \times 3}$;
    \item Image-encoding set: the feasible set ensures that the poisons respect the same encoding as the clean data $\mathcal{F}_{\mathcal{X}\text{img}}={[0..255]}^{H \times W \times 3}$;
    \item Neighborhood set: this is a subset of the previous one and is composed of all the images that are at a L1 norm of at most $\epsilon = \frac{32}{255}$ of an actual image $\mathcal{F}_{\mathcal{X}\text{nei}} = \left \{ x \in \mathcal{F}_{\mathcal{X}\text{img}} / \exists x_{a} \in D_{a} s.t. \left \| x - x_{a} \right \|_{1} \leq \epsilon \right \}$.
\end{itemize}
Figure \ref{fig:dp-avail-atk-test-acc-sgd-feasible-set} shows that the more constrained the feasible set, the less effective the resulting attack. It is to note, however, that early stopping helps in limiting the effects of the attack (see Figure \ref{fig:dp-avail-atk-best-test-acc-sgd-feasible-set} in the Appendix).

\paragraph{Neural Network architecture.} We perform the same attacks on the same learning pipelines but replace the CNN model with a Visual Transformer (ViT, \citet{dosovitskiy2020image}) tiny with patch size 8.
Since ViTs are predominantly trained using $\textsc{Adam}$-like approaches, we report the efficiency of our attacks using it.
Figure \ref{fig:dp-atk-vit8p} in the Appendix shows that ViT are also vulnerable to our data poisoning.
Although with far less success and with higher levels of contamination required.

%% file: sections/06-discussion.tex
\section{Concluding remarks}

In this work, we show that in training settings involving deep neural networks, which are not restricted to convex cases as in \cite{farhadkhani2022equivalence}, inverting malicious gradients can result in an effective data poisoning, achieving an availability attack.
Our threat model uses several assumptions that should be further explored in future work:
\begin{itemize}
    \item \textbf{The role of the auxiliary dataset}. In our experiments, it appears that the size of the auxiliary dataset plays only a small part in the success of the attack above a sufficient size. Further experiments should consider an auxiliary dataset with a distribution different than the training set or no auxiliary dataset at all.
    \item \textbf{Accessing the trained model's weights}. Performing an attack when the attacker estimates the victim's model with a surrogate model would open the threat model to a wider variety of cases.
    \item \textbf{The role of the feasible set}. Works on clean label poisoning \cite{geiping2021witches,shafahi2018poison} show that it is possible to design data poisons which deceive human annotators by resembling legitimate images. Future work should consider exploring more constraining feasible sets for the attacker to reach such level of stealthiness.
    \item Finally, while both our threat model and that of the closest work of \cite{farhadkhani2022equivalence}limit the fraction of poison that the attacker is allowed to inject each time, future work should explore data poisoning availability attacks limiting also the \textbf{number of times the attacker can craft its poisons} during the training phase.
\end{itemize}

We chose to solve the poisoning optimization problem with gradient-based approaches which are computationally intense, limiting us in experimenting with larger models and datasets.
Stronger gradient attacks that only require to steer the model in a direction only slightly dissimilar from honest gradients should also be explored to increase the chances for a stealth data poison to exist and to improve the attack success rate.
Stronger or less computationally expensive data poisoning approach (that may not rely on gradient attacks) can be experimented to enhance the attack success rate.

%% file: sections/07-conclusion.tex
While the tested defense mechanisms for the gradient case appear to generalize and defend against data poisoning, it has been shown that the latter can bypass certain defense mechanisms \cite{koh_stronger_2021} and that robust mean estimation only works up to a certain point \cite{elmhamdi2023impossible}.
This leaves room for potentially devastating data poisoning that can bypass defenses while mimicking an unstoppable gradient attack.

Our work asks whether or not gradient attacks were fundamentally more effective than data poisoning.
Our experiments yield a nuanced response.
On one hand, inverting malicious gradients sometimes result in a devastating data poisoning, and our results are the first to show the feasibility of a total availability attack on neural network via data poisoning.
On the other hand, the success rate of our attack with data poisoning had a lower success rate than their gradient attack counterparts.
The possibility of mimicking gradient attacks with feasible data poisoning should motivate further research in defense mechanisms and lower bounds to better assess the safety of machine learning algorithms in the presence of unreliable data.

\section{Broader Impact Statement}

This paper presents work whose goal is to advance the field of Robust Machine Learning. There are many potential societal consequences of our work, all of which fall under the usual considerations to take into account when considering tools that can facilitate data poisonning or make it more potent. In particular, our technique of inverting gradient' aims at showing that availability attacks -- which were believed to be specific to gradient attacks except for convex settings -- are doable with data poisonning alone, and in small amount. As such, this technique is of dual use and can be used (or is potentially already considered for use) by actors trying to further poison the data reservoirs used to train machine learning models.

%% file: sections/08-appendix.tex
\section*{Appendix}
\section{Discussion on realistic values of $\alpha$}

Machine Learning practitioners process data scraped online \cite{luccioni2021whats, schuhmann2022laion5b} or sent by users and trust their resulting models \cite{hoang2021tournesol}.
Since data poisoning has proven to be practical in real case scenario \cite{carlini2023poisoning}, we should treat models trained on scrapped or collaborative datasets with as much precaution as untrustworthy data.
A few redditors unpurposefully obtained a dedicated token in GPT-3's tokenizer by artificially inflating their online presence on the platform by massively posting over 160k posts on the ``r/counting'' subreddit on which people simply count \footnote{\href{https://www.lesswrong.com/posts/aPeJE8bSo6rAFoLqg/solidgoldmagikarp-plus-prompt-generation}{SolidGoldMagikarp (plus, prompt generation)}}.
Even worse, these data can potentially be sent by malevolent agents who can thus influence the models by legitimately participating to these data sources.

Estimating a realist value for $\alpha$ is difficult since the point for attackers is to not be detected.
Since bots represent a non-negligible part of online users (between 5\% \footnote{\url{https://twitter.com/paraga/status/1526237585441296385}} and 15\% \cite{varol_online_2017} on Twitter), we could expect the ratio of stealthy malevolent agents to be somewhat similar (if not higher).
``Armies'' as large as 2 million full-time individuals \cite{charon2021chinese} could be conducing campaigns on social media (each individual manually steering tens of social media accounts to escape automated-activity detection).
Collaborative datasets like Wikipedia \cite{carlini2023poisoning} could already be stealthily poisoned.
As such, we should not only presume that data we train on might have been poisoned, we should also consider the contamination ratio to be much higher than a fraction of a percent.
In this work, we considered a wide range up to the extreme case of $\alpha = 0.48$

\section{Complementary Figures and Tables}

\begin{table}[ht]
    \centering
    \caption{Architecture of the 1.6M parameters convolutional neural network used for our experiments.}
    \label{tab:cnn-arch}
    \begin{tabular}{cccc}
        \toprule
        Layer & \# of channels & kernel  & stride \\
        \midrule
        \texttt{Conv2d} & 32 & 5 $\times$ 5 & 2 \\
        \texttt{ReLU} &  &  & \\
        \texttt{Conv2d} & 64 & 5 $\times$ 5 & 2 \\
        \texttt{ReLU} &  &  & \\
        \texttt{Linear} & 512 &  & \\
        \texttt{ReLU} &  &  & \\
        \texttt{Linear} & 64 &  & \\
        \texttt{ReLU} &  &  & \\
        \texttt{Linear} & 10 &  & \\
        \bottomrule
    \end{tabular}
\end{table}

\begin{figure*}[ht]
    \centering
    \includegraphics[width=0.8\textwidth]{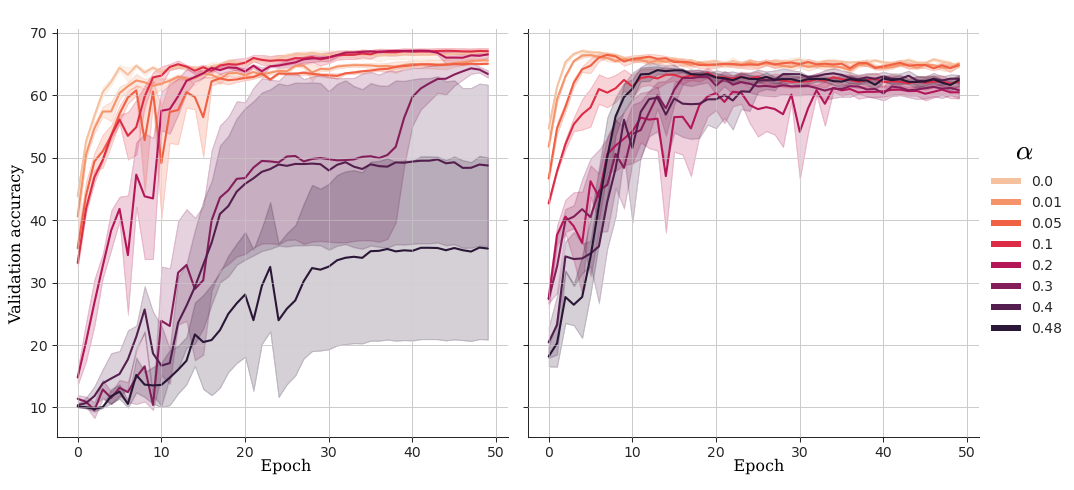}
    \caption{Validation accuracies of the CNN during training with the $\textsc{SGD}$ and $\textsc{Adam}$ update rule with the $\textsc{Average}$ aggregation function under the Gradient Ascent attack. This data poisoning manages to slow down the training but not degrade the model's performance to random levels. Error bars represent the standard error.}
    \label{fig:dp-tr-time-atk-val-acc-cosim-sgd-adam}
\end{figure*}

\begin{figure}[ht]
    \centering
    \includegraphics[width=\linewidth]{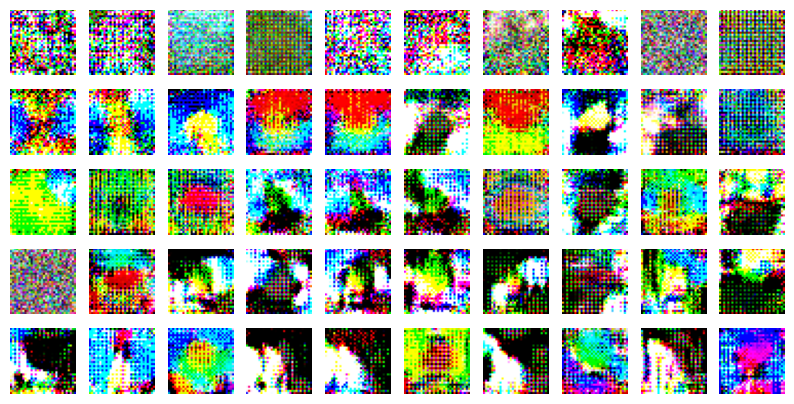}
    \caption{50 of the first 500 poisons crafted in the ($\textsc{SGD}$ \& $\textsc{Average}$, Little is Enough, $\alpha = 0.01$) setting.}
    \label{fig:dp-avail-sgd-little1e0-a001-50pois}
\end{figure}

\begin{figure*}[ht]
    \centering
    \begin{subfigure}{0.33\linewidth}
        \centering
        \includegraphics[width=\linewidth]{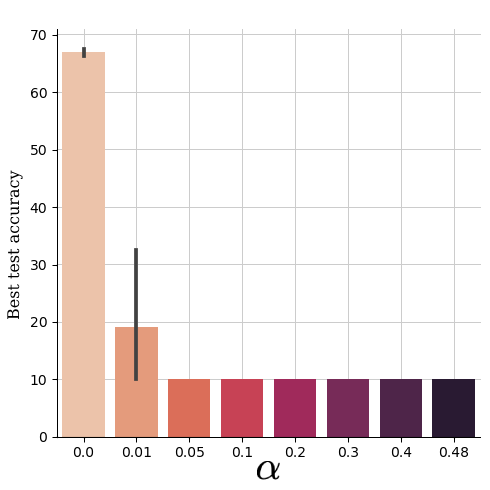}
        \caption{$\mathcal{F}_{\mathcal{X}\text{free}}$}
        \label{fig:dp-avail-atk-best-test-acc-sgd-free-set}
    \end{subfigure}
    \begin{subfigure}{0.33\linewidth}
        \centering
        \includegraphics[width=\linewidth]{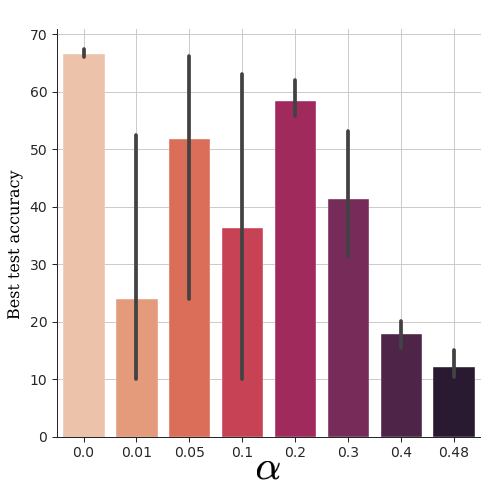}
        \caption{$\mathcal{F}_{\mathcal{X}\text{img}}$}
        \label{fig:dp-avail-atk-best-test-acc-sgd-encode-set}
    \end{subfigure}
    \begin{subfigure}{0.33\linewidth}
        \centering
        \includegraphics[width=\linewidth]{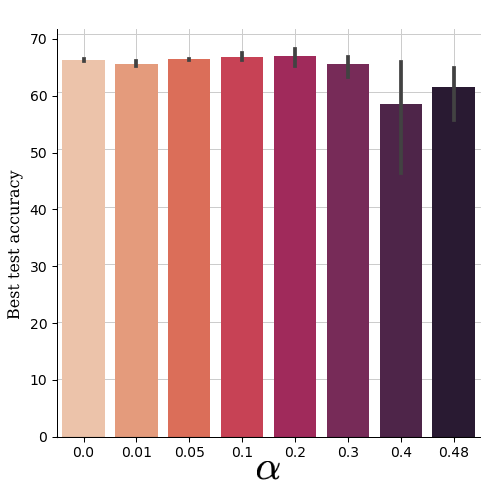}
        \caption{$\mathcal{F}_{\mathcal{X}\text{nei}}$}
        \label{fig:dp-avail-atk-best-test-acc-sgd-neighbor-set}
    \end{subfigure}
    \caption{Best test accuracies for the $\textsc{SGD};\textsc{Average}$ setting under the Little is Enough attack for different feasible sets.}
    \label{fig:dp-avail-atk-best-test-acc-sgd-feasible-set}
\end{figure*}

\begin{figure*}[ht]
    \centering
    \begin{subfigure}{0.62\linewidth}
        \centering
        \includegraphics[width=\linewidth]{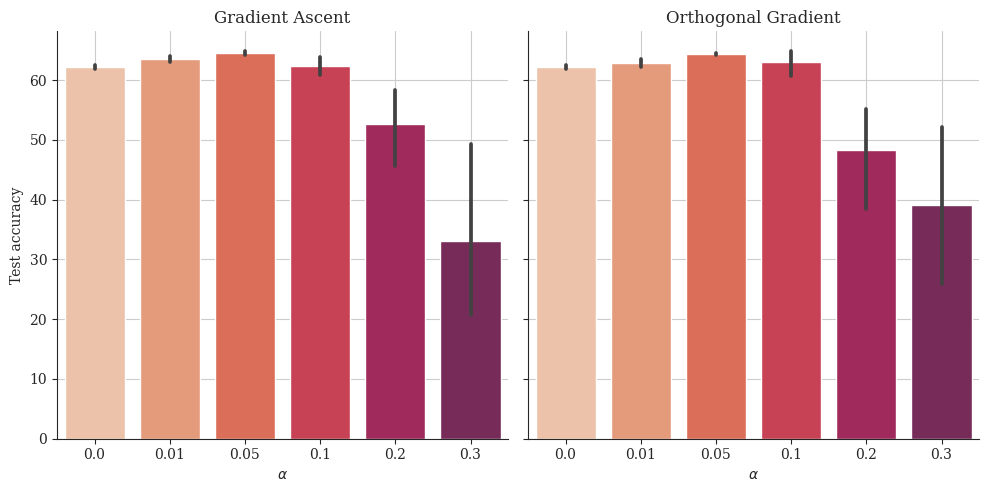}
        \caption{Availability attack.}
        \label{fig:dp-avail-atk-vit8p-best-test-acc-adam-cosim-ortho}
    \end{subfigure}
    \centering
    \begin{subfigure}{0.36\linewidth}
        \centering
        \includegraphics[width=\linewidth]{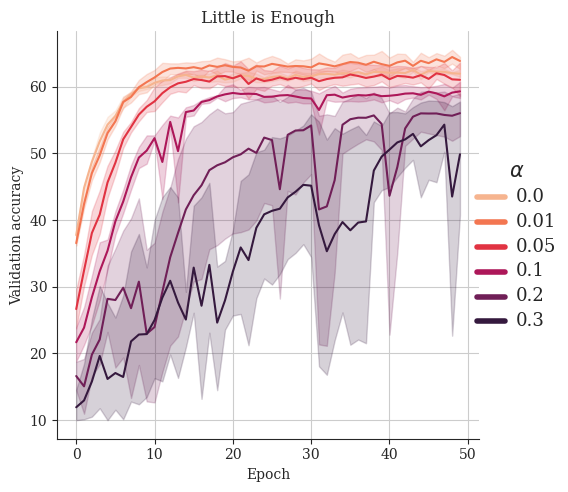}
        \caption{Slow down attack.}
        \label{fig:dp-tr-time-atk-vit8p-val-acc-adam-little}
    \end{subfigure}
    \caption{Visual Transformer (ViT) tiny with patch size 8 under different attacks. Little is Enough performs a slow down attack whereas Gradient Ascent and Orthogonal Gradient are able to perform an availability attack for high enough contamination levels $\alpha$.}
    \label{fig:dp-atk-vit8p}
\end{figure*}